\documentclass[10pt,twocolumn,letterpaper]{article}

\usepackage{cvpr}              

\usepackage{graphicx}
\usepackage{amsmath}
\usepackage{amssymb}
\usepackage{booktabs}
\usepackage{physics}
\usepackage{tabularx}
\usepackage[pagebackref,breaklinks,colorlinks]{hyperref}
\usepackage[table]{xcolor}

\usepackage{bbding}
\usepackage[capitalize]{cleveref}
\crefname{section}{Sec.}{Secs.}
\Crefname{section}{Section}{Sections}
\Crefname{table}{Table}{Tables}
\crefname{table}{Tab.}{Tabs.}
\newcommand{\Tref}[1]{Table~\ref{#1}}

\newcommand{\Fref}[1]{Fig.~\ref{#1}}
\newcommand{\Sref}[1]{Section~\ref{#1}}
\newcommand{\bm}[1]{{\mbox{\boldmath $#1$}}}
\def\cvprPaperID{1453} 
\def\confName{CVPR}
\def\confYear{2022}

\begin{document}

\title{Universal Photometric Stereo Network using Global Lighting Contexts}
\author{Satoshi Ikehata\\
National Institute of Informatics (NII)\\
{\tt\small sikehata@nii.ac.jp}
}
\twocolumn[{%
\renewcommand\twocolumn[1][]{#1}%
\maketitle
\vspace{-30pt}
\begin{center}
    \centering
    \captionsetup{type=figure}
    \includegraphics[width=175mm]{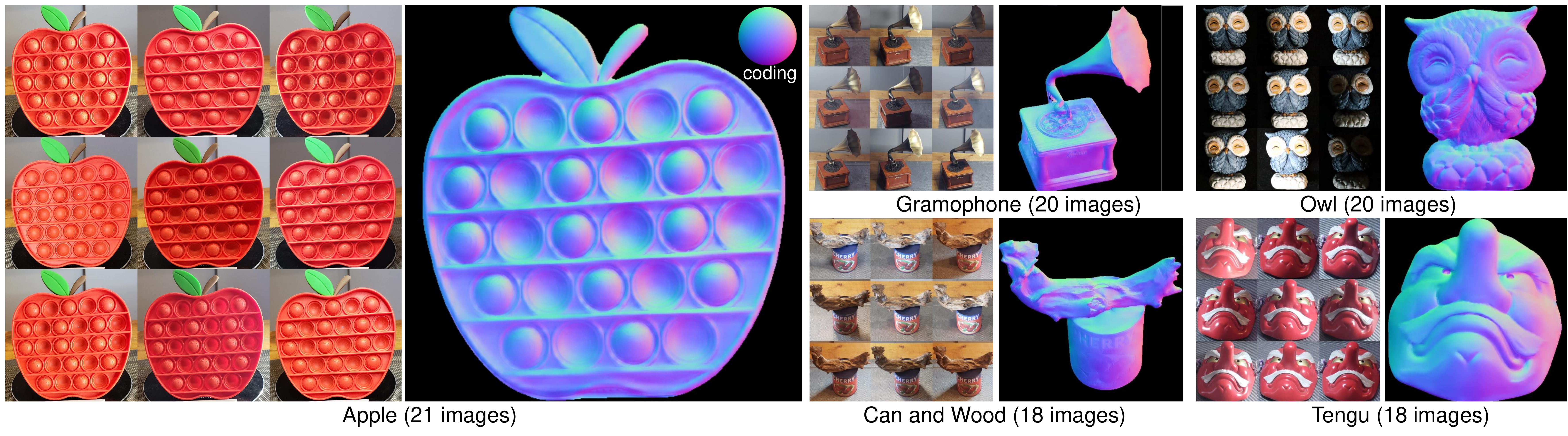}
    \vspace{-20pt}
    \captionof{figure}{Given images of an object under different lighting conditions, our method can recover the detailed surface normal map. Images were captured by a smartphone camera with moving handheld lights under the indoor natural illumination. Please see details in~\Sref{sec:realdata}.}
    \label{fig:realdata}
\end{center}%
}]
\begin{abstract}
This paper tackles a new photometric stereo task, named universal photometric stereo. Unlike existing tasks that assumed specific physical lighting models; hence, drastically limited their usability, a solution algorithm of this task is supposed to work for objects with diverse shapes and materials under arbitrary lighting variations without assuming any specific models. To solve this extremely challenging task, we present a purely data-driven method, which eliminates the prior assumption of lighting by replacing the recovery of physical lighting parameters with the extraction of the generic lighting representation, named global lighting contexts. We use them like lighting parameters in a calibrated photometric stereo network to recover surface normal vectors pixelwisely. To adapt our network to a wide variety of shapes, materials and lightings, it is trained on a new synthetic dataset which simulates the appearance of objects in the wild. Our method is compared with other state-of-the-art uncalibrated photometric stereo methods on our test data to demonstrate the significance of our method.
\end{abstract}
\vspace{-20pt}
\section{Introduction}
\label{sec:intro}
Photometric stereo is a problem of recovering the surface normal map from appearances of an object under varying lighting conditions. For decades, a broad spectrum of techniques have been proposed to expand the scope of target geometry, material and acquisition setup~\cite{Goldman2005,Shi2014,Ikehata2012,Ikehata2014b} in the framework of the physics-based inverse rendering. Recently, advances in deep learning have been eliminating the dependence on physics-based modeling from photometric stereo methods, which contributes to handle complex optical phenomena which are hardly described in a mathematically tractable form\cite{Santo2017,Taniai2018,Ikehata2018,Chen2020}. 

However, despite the long journey in this research field, each photometric stereo algorithm is still limited to a specific physical lighting model, which severely compromises its usability. In reality, most recent (semi-)calibrated~\cite{Santo2017,Ikehata2018,Chen2018,Taniai2018,Santo2020} and uncalibrated~\cite{Chen2020,Kaya2021} photometric stereo methods still assume the single lighting in a dark environment. Others address natural lighting conditions, however both calibrated and uncalibrated methods still assume convex Lambertian surfaces and their lighting models are limited to spherical harmonics lighting~\cite{Basri2007,Haefner2019}, dominant sun lighting~\cite{Ackermann2012,Hold2019} and equivalent directional lighting~\cite{Mo2018,Guo2021} which cannot represent the complex illumination.

To address this limitation, this paper proposes the ``third'' task in photometric stereo problem following calibrated and uncalibrated tasks. We name it {\it universal} photometric stereo (UniPS) which denotes the setup without prior assumption of physical lighting models; hence, arbitrary lighting conditions should be considered unlike calibrated and uncalibrated tasks that consider specific ones as in~\Tref{table:tasks}.\footnote{It would be ideal if the task were universal on materials as well. However, since some objects such as mirrors and transparent objects must be excluded, we consider only lighting conditions to be universal in this task.}

In this paper, we present a first viable method for UniPS based on key insights as follows. Conventional uncalibrated photometric stereo algorithms recovered physical lighting parameters and surface normals sequentially or simultaneously; thus, constrained by specific lighting models. However, we show that the recovery of physical lighting parameters is {\it not} essential in a UniPS network, and can be replaced with the extraction of {\it global lighting contexts} from individual images by their interaction with others. In order to keep the receptive field of our network constant, contexts are extracted at the predefined {\it canonical resolution} which is independent of the input image resolution. By using global lighting contexts like calibrated lighting parameters, surface normal vectors can be recovered in similar to existing pixelwise calibrated photometric stereo networks (\eg \cite{Ikehata2018,Ikehata2021}) which easily scale to high-resolution images. 
\begin{table}[t]
\newcolumntype{C}{>{\centering\arraybackslash}X}
\setlength{\tabcolsep}{2.0mm} 
    \centering
    \vspace{-5pt}
    \caption{Comparison of different photometric stereo tasks.}
    \vspace{-10pt}
    \small
    {\renewcommand\arraystretch{1.0}
    \begin{tabularx}{80mm}{XCCC}
    \toprule
         & Lighting Calibration & Lighting Model & Lighting Condition\\
        \midrule
        Calibrated & Required & Required & Specific\\
        Uncalibrated & Free  & Required & Specific\\
        \cellcolor{green!5}Universal & \cellcolor{green!5}Free  & \cellcolor{green!5}Free & \cellcolor{green!5}Arbitrary \\
    \bottomrule
    \end{tabularx}
    }
    \label{table:tasks}
    \vspace{-10pt}
\end{table}
While our network drops a prior assumption of lighting, its adaptation to diverse shapes, materials and lightings has to be ensured by training data. Since existing training datasets for photometric stereo networks are limited to the single, directional lighting setup, we create a dataset for our task by physically rendering the appearance of objects with more than $10,000$ combinations of shape, material and lighting using high quality commercial 3-D assets. We also create an evaluation dataset with $50$ sets of attributes using different assets to compare our method with state-of-the-art uncalibrated photometric stereo algorithms specifically designed for directional lighting~\cite{Chen2020} and natural lighting~\cite{Mo2018,Haefner2019}. Finally, the qualitative evaluation demonstrates that our method even works for real objects under the challenging spatially-varying lighting conditions that were conventionally considered to be intractable (See~\Fref{fig:realdata}).
\section{Related Work}
Here we briefly review uncalibrated photometric stereo methods under the directional or natural lighting. Due to the space limit, other categories (\eg calibrated setup~\cite{Ikehata2012,Ikehata2014a, Ikehata2014b,Ikehata2018, Ikehata2021}, near-light setup~\cite{Iwahori1990,Santo2020} or multi-view setup~\cite{Park2013,Li2020ps}) are not included in this survey. However, we emphasize that all the methods more or less assume any of physical lighting models; hence, are limited to specific lighting setups.
\vspace{4.0pt}\\
\noindent\textbf{Uncalibrated, Directional Lighting:} Since Woodham~\cite{Woodham1980} presented the first Lambertian photometric stereo algorithm in 1980, methods following its setup have been called calibrated photometric stereo which assumes an orthographic camera and a known single directional lighting. The uncalibrated task is almost identical to the calibrated one except that lighting parameters are unknown. Until very recently, most uncalibrated photometric stereo algorithms assumed Lambertian integrable surfaces and the goal was to resolve the General Bas-Relief ambiguity~\cite{Hayakawa1994} between geometry and light. Various cues were employed for resolving this ambiguity, which include inter-reflections~\cite{Drobohlav2002}, entropy of surface albedo~\cite{Alldrin2007a}, color profile~\cite{Shi2010}, diffuse maxima~\cite{Favaro2012}, reflectance symmetry~\cite{Wu2013} and perspective geometry~\cite{Papadhimitri2013}. Though there were very few methods for non-Lambertian surfaces due to their ill-posed nature, Lu~\etal~\cite{Feng2013} utilized the statistical distribution of the intensity profile to recover isotropic non-Lambertian surfaces. 

In 2018, the first deep neural network for uncalibrated photometric stereo was presented~\cite{Chen2018}. However, it simply dropped the lighting channel from its calibrated variant presented at the same time and the performance was quite limited. Hence, authors extended their work to the two-step approach~\cite{Chen2019} where only lighting information was firstly recovered, and then it was used as input of the calibrated photometric stereo network. Later, authors further updated their work by feeding the surface normal estimation result back to the lighting prediction to improve its accuracy~\cite{Chen2020}. Building upon this work, Kaya~\etal~\cite{Kaya2021} have recently utilized the lighting prediction result from~\cite{Chen2020} to recover surface normals in the neural inverse rendering framework.
\vspace{4.0pt}\\
\noindent\textbf{Uncalibrated, Natural Lighting:} Although most photometric stereo methods assume the single directional lighting condition, some literature exists which addresses natural lighting conditions. However, inversely decomposing the natural light reflected on non-Lambertian surfaces is intractable in the inverse rendering framework, so it is common to assume convex Lambertian surfaces and approximated lighting models~\cite{Basri2007,Mo2018,Haefner2019,Brahimi2020,Guo2021}. 

The first uncalibrated photometric stereo algorithm under natural lighting was proposed by Barsri~\etal~\cite{Basri2007} which approximated natural lighting as global first-order spherical harmonics. Though a global concave-convex ambiguity exists in estimated surface normals due to the lighting approximation with an orthographic camera, Brahimi~\etal~\cite{Brahimi2020} recently proved that the perspective
integrability constraint makes the problem well-posed. Mo~\etal~\cite{Mo2018} presented another uncalibrated photometric stereo algorithm under natural lighting. They proposed the equivalent directional lighting model which decomposed the entire task into the patch-wise directional uncalibrated photometric stereo problem. 
This work was later extended by authors where MRF-based global optimization and rotation averaging were introduced for a better normal patch integration~\cite{Guo2021}.
These methods worked under an orthographic camera, however when solving the global orthogonal ambiguity with integrability, there was a binary ambiguity left which requires to be solved manually and a non-integrable surface can not be recovered in theory. To address this issue, Haefner~\etal~\cite{Haefner2019} presented a variational method which directly recovered a depth map rather than a surface normal map to handle non-integrable surfaces. However, this method requires initial geometry which is basically recovered from the object outline and its performance drastically affected by its quality. 
\section{Universal Photometric Stereo}
\begin{figure}[!t]
	\begin{center}
		\includegraphics[width=80mm]{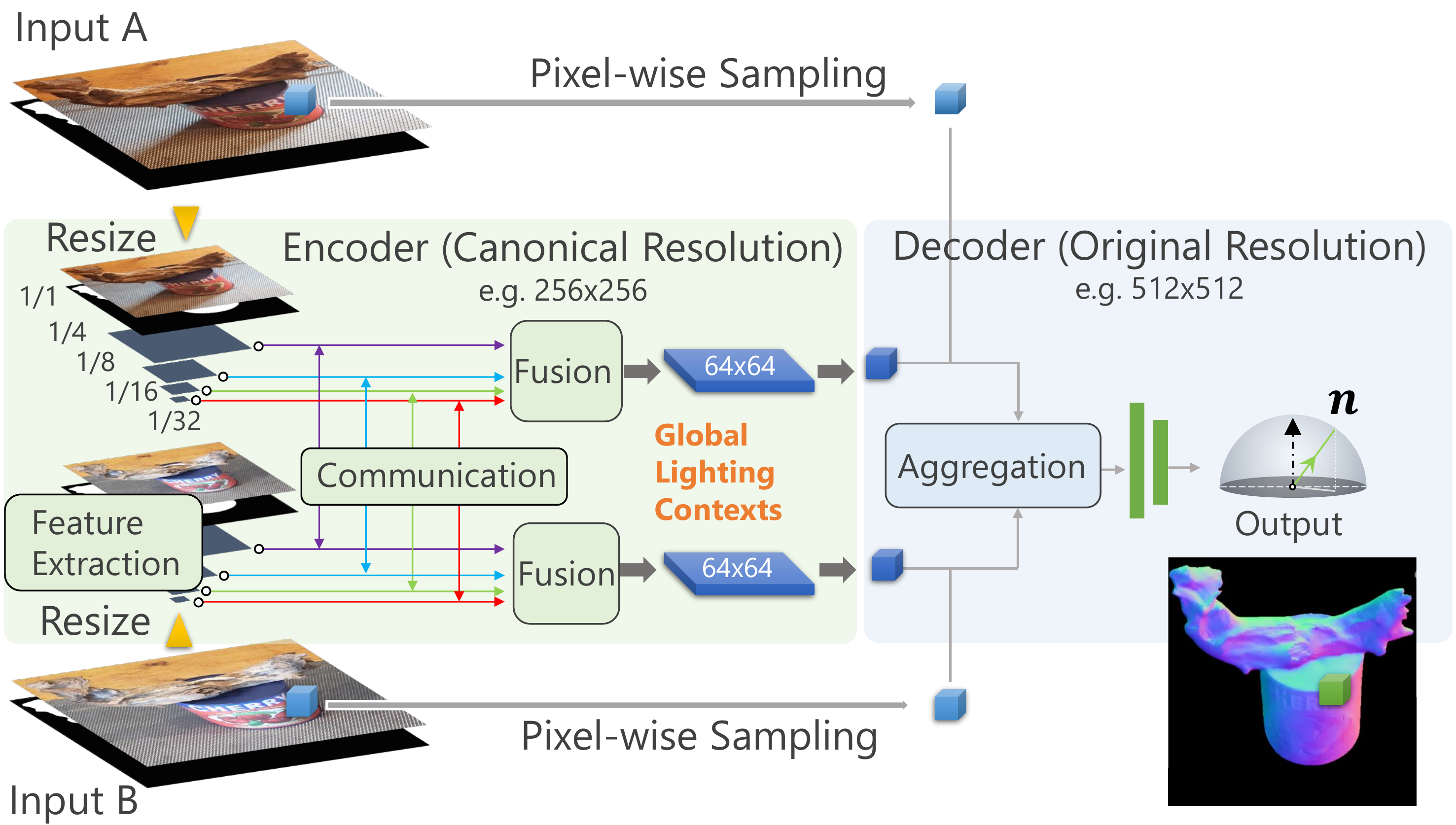}
	\end{center}
	\caption{Illustration of the proposed method. Given arbitrary number of image crops (\eg two in this figure) of an object under different lighting conditions, they are resized to the {\it canonical resolution} and passed to the encoder to extract global lighting contexts. Then, for each pixel in the {\it original image}, raw image values and an interpolated global lighting context are concatenated to be fed to the decoder that aggregates features on the lighting axis and recovers the surface normal vector at the location.}
	\vspace{-5pt}
	\label{fig:architecture}
\end{figure}
In this section, we preliminarily define the UniPS problem setup. Given a set of $q$ RGB images $I_{\{1,\dots,q\}}$ captured under unknown, arbitrarily varying lighting conditions with a fixed camera, and a binary mask $M$ to specify the target object in the image, our goal is to recover the unit surface normal vector $\vb*{n}$ at pixels which belong to the object. The object mask is friendly provided using any existing foreground extraction methods such as~\cite{Sofiiuk2021}. UniPS doesn't put any particular constraints on the camera model, however an orthographic, linear camera is practically assumed in this work as training images are rendered by the camera.

The main difficulty in UniPS is due to the lack of prior knowledge of lighting which disables most of existing photometric stereo algorithms designed under specific physical lighting models. In UniPS, lighting conditions could include near or distant, directional, point, area, natural or even mixture of them (\eg putting active near area lights under passive natural light as demonstrated in~\Fref{fig:realdata}). It is highly possible that the lighting is {\it spatially-varying} which cannot be represented by a global lighting model. This may be the case, for example, in most indoor natural scenes~\cite{Li2020}. In a general context, even inter-reflections and cast shadows can also be a part of spatially-varying lightings. 

Since the ultimate goal of UniPS is to realize a truly practical photometric stereo method, the available geometries and materials (except for extreme cases such as transparent and perfect mirror objects) should be diverse as well and the number of input images and their resolution should be arbitrary. Although accepting arbitrary number of input images was considered as an important requirement in recent works (\eg \cite{Taniai2018,Ikehata2018,Chen2018}), there are still some methods that do not meet this requirement~(\eg \cite{Lichy2021}). Scalability is basically not a problem in pixel-by-pixel algorithms (\eg \cite{Ikehata2012,Ikehata2018}), but it often becomes a major issue in methods that  use the entire image information such as convolutional neural networks and global optimization (\eg \cite{Mo2018,Chen2020}). 
\section{Method}
\subsection{UniPS Network with Global Lighting Contexts}
This paper presents a first viable universal photometric stereo network. As illustrated in~\Fref{fig:architecture}, our network consists of an encoder and decoder. The encoder extracts {\it global lighting contexts} from images and an object mask, which is a generic lighting representation that corresponds to physical lighting parameters (\eg light direction) in deep uncalibrated photometric stereo networks~\cite{Chen2019, Chen2020}. The decoder takes all raw image values and the interpolated global lighting context at each pixel and predict its surface normal.

Our network architecture has two major differences from basic encoder-decoder architectures that take a single image as input. First, our network takes multiple images as input; hence, features must be embedded in the latent space considering the interaction of them. Therefore, we perform feature communication in the encoder and aggregation in the decoder, respectively. Second, unlike typical architectures where encoded features are directly passed to the decoder, we use different working resolutions for the encoder and decoder. The working resolution for the decoder is same as the original image resolution, but the encoder takes as input images that have been resized to the pre-defined {\it canonical resolution}, which is basically smaller than the original resolution and its output is passed to the decoder after inversely converted for the decoder's working resolution. 

There are two major advantages to use different working resolutions. First, the scalability to the image size is ensured because memory requirements for the encoder depend only on the canonical resolution, not on the original one, while the decoder processes each pixel one by one. The second and more important reason is to keep the receptive field of the encoder invariant to the input image size. Without this, the networks' receptive field may not cover the entire object in extremely high-resolution test images.
\subsection{Framework Components}
\noindent\textbf{Preprocessing:} The pixel value range of individual images can significantly vary under different lightings. Therefore, we divide each image by its mean for normalization. Note that the common zero-mean normalization is not used to avoid emphasizing the area of low signal-to-noise ratio. We then crop the rectangular object bounding region with a small margin (\eg four pixels) based on the object mask to confirm that the object is placed in the middle of the crop and covered from edge to edge. 
\vspace{4.0pt}\\
\noindent\textbf{Encoder:}
Preprocessed crops of the images and mask are bilinearly resized to the predefined $s\times s$ canonical resolution. They are then concatenated $C_{\{1,\dots,q\}}\in\mathbb{R}^{s\times s\times (3+1)}$ and passed to the encoder. The output of the encoder is embedded features for individual images whose size is a quarter of the canonical resolution $\mathcal{G}_{\{1,\dots,q\}}\in \mathbb{R}^{\frac{s}{4}\times \frac{s}{4} \times d_e}$ where $d_e$ is the embed dimension. We call them as global lighting contexts because the difference of features among images should only be attributed to the difference of lighting conditions. Global lighting contexts are also an analogy of physical lighting parameters similarly recovered by the conventional uncalibrated photometric stereo methods. However, unlike physical parameters were basically assigned to each image globally (\eg light direction~\cite{Chen2019}, spherical harmonics~\cite{Haefner2019}), a unique global lighting context is assigned at each location (therefore we name context`s'); hence has the capacity to represent spatially-varying phenomena such as near-lighting, inter-reflections and cast shadows.
\begin{figure}[!t]
	\begin{center}
		\includegraphics[width=80mm]{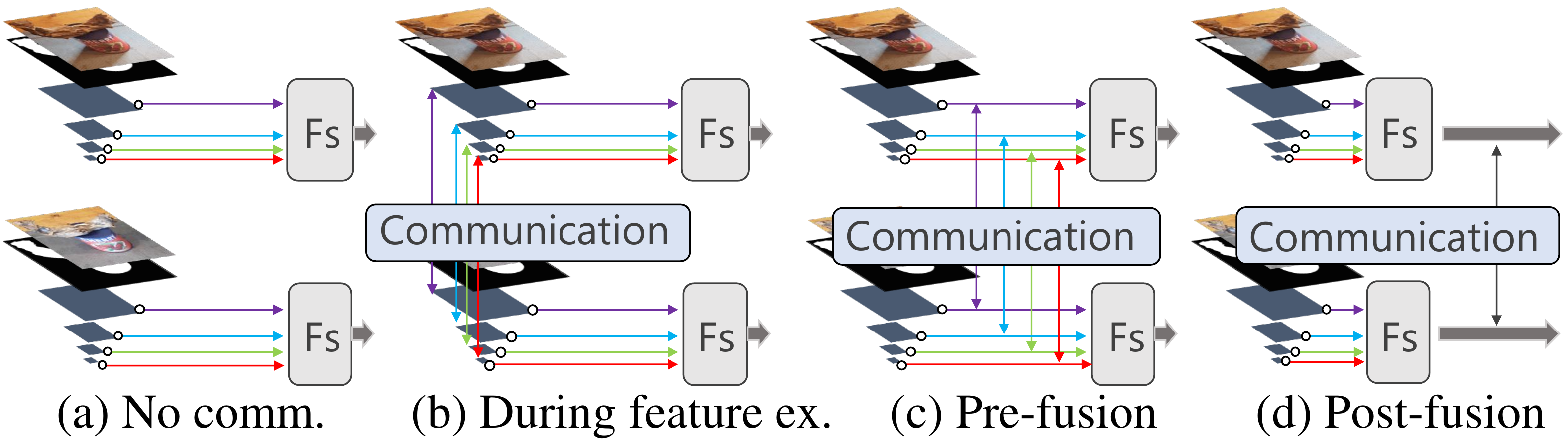}
	\end{center}
	\vspace{-10pt}
	\caption{The illustration of four different variants of the encoder with regard to the timing of the inter-image communication.}
	\vspace{-10pt}
	\label{fig:interaction}
\end{figure}

The encoder mutually embeds global lighting contexts via the image-wise feature extraction and inter-image feature communication. The former extracts multi-scale feature maps using a backbone (\eg SwinTransformer~\cite{Liu2021Swin}) followed by the basic multi-scale feature fusion~\cite{Xiao2018}. The latter propagates features across images without changing the feature dimension by pixelwisely employing a single Transformer layer~\cite{Vaswani2017} as with the recent calibrated photometric stereo network~\cite{Ikehata2021}. As illustrated in~\Fref{fig:interaction}, there are multiple feasible designs of the encoder depending on how information propagate via inter-/intra-image interactions. The best design of them will be discussed in the ablation study in~\Sref{sec:ablation}. As it will turn out, the feature communication should be conducted just before the multi-scale feature fusion (\ie pre-fusion). Please refer to the supplementary for more details about individual implementations of the feature extraction and communication.
\vspace{4.0pt}\\
\noindent\textbf{Decoder:} For each coordinates $\bm{x}$ of the {\it original} resolution, the decoder inputs a set of image values $I_{\{1,\dots,q\}}(\bm{x})$ and a global lighting context $\mathcal{G}_{\{1,\dots,q\}}(\mathcal{S}(\bm{x}))$, and outputs the unit normal vector at the location. Here, $\mathcal{S}$ is the sampling operation to fetch the value of the global lighting context corresponding to $\bm{x}$ using bilinear interpolation. Note that avoiding the resize of global lighting contexts to the original resolution also contributes to the scalability. 

The decoder is composed of the feature aggregation and the surface normal recovery. The former further propagates information across different lighting conditions with extra Transformer layers, and then squeezes the lighting channel with the pooling by multi-head attention (PMA)~\cite{Lee2019}. The output of PMA is a pixelwise feature vector independent of number of input images. The latter feeds the feature to the multi-layer perceptron with one hidden layer to output the surface normal vector. We should note that when no communication is conducted in the encoder as in~\Fref{fig:interaction}, it can be deemed that global lighting contexts have been extracted at the aggregation step in the decoder. 
\vspace{4.0pt}\\
\noindent\textbf{Training Loss:} The mean squared error between ground truth and predicted surface normals is used as the training loss. Because of the efficiency, we compute the loss only on samples from specific locations, not from the entire image. For each object, we uniformly sample pixels at the original resolution whose centers can be projected exactly onto centers of pixels at the canonical resolution. Then, we further sample pixels from a limited number of random locations (\eg 2500) at the original resolution which allows for the normal estimation with a sub-pixel global lighting context.
\vspace{4.0pt}\\
\noindent\textbf{Network Analysis}: Our method is data-driven; hence satisfies the requirements of UniPS. This is achieved by global lighting contexts, which eliminate the need for the physical lighting parameter recovery and makes our method feasible to handle complex spatially-varying lighting effects.

Our framework is scalable by introducing the canonical resolution for encoding global lighting contexts and pixelwisely applying the decoder. The network's receptive field is independent of the input image size as well as there is no upper limit to the test image size. By utilizing the self-attention mechanism for the feature interaction across images, the network is applicable to an arbitrary number of images as long as the computational resource permits. Given this capacity, the availability of varying shapes, materials and lightings are secured by training data.
\subsection{PS-Wild Training Dataset}
We need sufficient training examples that properly characterize the universal photometric stereo task. Though two major synthetic photometric stereo datasets (\ie Blobby and Sculpture~\cite{Chen2018}, CyclesPS~\cite{Ikehata2018}) have been presented to date, they were for the methods under the single directional lighting and not applicable to UniPS. Therefore, we present a new photometric stereo dataset, named {\it PS-Wild}, which simulates the appearance of objects with diverse geometries and materials captured under various lighting conditions in the wild. The idea is simply to ask a physically-based renderer (\ie Blender Cycles~\cite{Blender}) to synthesize a large number of images under general lightings, taking full advantage of high quality commercial 3-D assets. 

To assemble an appropriate collection of object appearances for our PS-Wild dataset, we browsed through online marketplaces looking for 3-D assets that satisfied three main desiderata. First, we should use 3-D models that have sufficient complexity. We don't want the objects whose surface normal distribution deviates from ones in the wild (\eg too many planar objects or low-poly models are not adequate). Second, we want 3-D materials to be as diverse as possible. In our problem, we don't assume specific materials as well as lighting models, so the training data needs to be chosen to cover diverse materials in the wild. In addition, textures need to be realistic. In the existing synthetic datasets, the entire image was rendered with a single BRDF~\cite{Chen2018}, or pixelwisely different random BRDFs~\cite{Ikehata2018,Logothetis2021}, but the actual surface texture has some realistic rules as mentioned in previous studies~\cite{Goldman2005,Alldrin2007a}. Hence, it is worth using texture maps that were designed by professionals. Third, we want to render scenes under diverse lighting conditions. In this regard, the most practical method we found is to use the HDR image-based (HDRI) lighting that covers various indoor and outdoor scenes. To add more details, this method uses an HDR environment map to place an omnidirectional light source. Each ray from the light is self-shielded before it hits the surface, so a variety of spatially-varying lighting effects (\eg cast shadows, near-lighting) are baked. 

Based on these desiderata, we choose Adobe Stock~\cite{AdobeStock} as our assets to create the training dataset. This collection consists of over 17,000 3-D assets including 3-D models, materials (texture maps) and lightings (environment maps). From all available assets, we downloaded 410 3-D models, 926 materials and 31 lightings. Most of data in Adobe Stock assets are 3-D models and we actually used all the materials and lightings available at the time we accessed there (Oct. 2021). The number of assets itself may not seem very large, but in reality, a countless number of images can be synthesized by various augmentation techniques such as rotation and color transformation.

After acquiring 3-D assets, we apply our computational pipeline to generate images with ground truth surface normal maps. For each 3-D object in assets, we randomly assign one material and one environment map. Then we apply several random rotations of the object until the Shannon entropy of its surface normal distribution becomes more than a threshold (\ie 4.0 in our method) and if the maximum entropy is less than the threshold, the 3-D model is discarded from assets. We then scale the object to make sure that the object's outline reaches the edge of the image. We render $10$ of $512\times 512$, $16$-bit images per object. For each rendering, we randomly rotate the environment map on the spherical axis to make the variation of lighting conditions and automatically adjust the exposure of the camera to make the dynamic range of rendered images consistent. We turn on the ray tracer to render the cast shadow and inter-reflection for adding the spatially-varying lighting effects. Finally, we got 10,099 objects with a different pose, material and lighting. 
\subsection{PS-Wild Test Dataset}
\begin{table*}[!t]
\begin{minipage}[t]{80mm}
\newcolumntype{C}{>{\centering\arraybackslash}X}
\setlength{\tabcolsep}{0.8mm} 
    \centering
    \caption{Comparison with different image feature extractors.}
    \vspace{-10pt}
    \small
    {\renewcommand\arraystretch{0.7}
    \begin{tabularx}{80mm}{XCCCC}
    \toprule
         & SwinT~\cite{Liu2021Swin} & ViL\cite{Zhang2021} & ResNet101 & ResNet50 \\
         \midrule
        Directional & \textbf{19.7} & 29.5 & 48.6 & 50.6 \\
        HDRI & \textbf{16.8} & 23.2 & 45.8 & 44.0 \\
        Dir.+HDRI & \textbf{16.1} & 24.0 & 45.2 & 45.0 \\
    \bottomrule
    \end{tabularx}
    }
    \label{tab:enc}
\end{minipage}
\hspace{10mm}
\begin{minipage}[t]{80mm}
\newcolumntype{C}{>{\centering\arraybackslash}X}
\setlength{\tabcolsep}{1.0mm} 
    \centering
    \caption{Comparison with different encoder designs.}
    \vspace{-10pt}
    \small
    {\renewcommand\arraystretch{0.7}
    \begin{tabularx}{80mm}{XCCCC}
    \toprule
         & Base & Dur-Ext. & Pre-Fus. & Post-Fus. \\
        \midrule
        Directional & 19.7 & 22.5 & \textbf{17.0} & 25.6 \\
        HDRI & 16.8 & 20.1 & \textbf{14.5} & 22.5 \\
        Dir.+HDRI & 16.1 & 19.4 & \textbf{13.8} & 21.1 \\
    \bottomrule
    \end{tabularx}
    }
    \label{tab:interaction}
\end{minipage}
\end{table*}

\begin{table*}[t]
\begin{minipage}[t]{80mm}
\newcolumntype{C}{>{\centering\arraybackslash}X}
\setlength{\tabcolsep}{2.0mm} 
    \centering
    \vspace{-5pt}
    \caption{Comparison with different canonical resolutions.}
    \vspace{-10pt}
    \small
    {\renewcommand\arraystretch{0.7}
    \begin{tabularx}{80mm}{XCCC}
    \toprule
         & $128\times 128$ & $256\times 256$ & $512\times 512$ \\
        \midrule
        Directional & 19.9 & \textbf{19.7} & N.A \\
        HDRI & 18.1 & \textbf{16.8} & N.A \\
        Dir.+HDRI & 17.0 & \textbf{16.1} & N.A \\
    \bottomrule
    \end{tabularx}
    }
    \label{tab:cano}
\end{minipage}
\hspace{10mm}
\begin{minipage}[t]{80mm}
\newcolumntype{C}{>{\centering\arraybackslash}X}
\setlength{\tabcolsep}{2mm} 
    \centering
    \vspace{-5pt}
    \caption{Spatially-varying vs uniform global lighting contexts.}
    \vspace{-10pt}
    \small
    {\renewcommand\arraystretch{0.7}
    \begin{tabularx}{80mm}{lCC}
    \toprule
         & Spatially-varying & Spatially-uniform \\
        \midrule
        Directional & \textbf{19.7} & 40.5 (trival) \\
        HDRI & \textbf{16.8} & 40.5 (trivial) \\
        Dir.+HDRI & \textbf{16.1} & 40.5 (trivial) \\
    \bottomrule
    \end{tabularx}
    }
    \label{tab:global}
\end{minipage}
\end{table*}

\begin{table}
\newcolumntype{C}{>{\centering\arraybackslash}X}
\setlength{\tabcolsep}{1mm} 
    \centering
    \caption{Comparison with different feature aggregation methods.}
    \vspace{-10pt}
    \small
    {\renewcommand\arraystretch{0.7}
    \begin{tabularx}{80mm}{XCCC}
    \toprule
         & Max-pool & TF+PMA (3L) & TF+PMA (6L) \\
        \midrule
        Directional & 59.3 & \textbf{19.7} & 37.5 \\
        HDRI & 39.3 & \textbf{16.8} & 29.4 \\
        Dir.+HDRI & 39.0 & \textbf{16.1} & 30.6 \\
    \bottomrule
    \end{tabularx}
    }
    \label{tab:dec}
\end{table}
We also create a test dataset for the evaluation purpose. The computational pipeline of generating images is same as one for the training dataset but different 3-D assets are used for the fair evaluation; 25 objects from CGTrader~\cite{CGTrader}, 50 materials from ShareTextures~\cite{ShareTextures} and 50 environment maps from sIBL Archive~\cite{sIBLArchive}. For each 3-D model, we assign two sets of a material and environment map resulting in 50 different sets of the object, material and environment map. Unlike training dataset, we carefully pick six textures per each texture category; Concrete, Fabric, Floor, Ground, Wood and Metal as categorized in ShareTextures. In order to properly evaluate the performance of a method under various lighting conditions, we render images for the same set of the object and material using three different lighting methods; (a) single directional lighting (uniformly sampled), (b) HDRI lighting (same as training) and (c) mixture of (a) and (b). The image resolution is also $512\times 512$ but the number of images is $32$ to evaluate the performance with varying number of input images. 
\vspace{4.0pt}\\
\noindent\textbf{Dataset Analysis:} Our test dataset characterizes the universal photometric stereo task. Objects with both convex and non-convex geometries and a variety of spatially-varying materials involving diffuse, specular and {\it metallic} are rendered with three different lighting methods including the HDRI lighting which exhibits challenging spatially-varying lighting effects. Therefore, methods that assume a specific physical lighting model are not suitable for this dataset. 
\section{Results}
We conduct experiments on synthetic and real test data. We first ablate the important design elements of our architecture and then we compare the proposed universal photometric stereo network with the previous state-of-the-arts on the uncalibrated photometric stereo task~\cite{Chen2019,Mo2018,Haefner2019}. For the convenience, we henceforth refer to the proposed method as UniPS-GLC (Universal Photoemtric Stereo network using Global Lighting Contexts) as necessary. 
\vspace{4.0pt}\\
\noindent\textbf{Training Details:}
Our network was trained from scratch on a NVIDIA Quadro RTX 8000 machine with AdamW~\cite{Loshchilov2019} optimizer for 20 epochs using a step decay learning rate schedule ($\times 0.8$ every three epochs). A batch size of 3, an initial learning rate of 0.0001, and a weight decay of 0.05 were used. The number of random samples from the original resolution was fixed by $2500$. In total, it took roughly 48 hours to train the network for each configuration.
\vspace{4.0pt}\\
\noindent\textbf{Inference Time:} The inference time of our method depends on the number and resolution of input images. In the case of $32$ of $512\times 512$ images as input, it takes less than a few seconds excluding IO on GPU. This is slightly slower than the simpler deep photometric stereo networks (\eg $0.5$ sec in~\cite{Chen2019}), however more than hundred times efficient than inverse rendering based methods~\cite{Mo2018,Taniai2018,Haefner2019,Kaya2021}.
\vspace{4.0pt}\\
\noindent\textbf{Evaluation metric:} Evaluation is based on the mean angular errors (MAE) between predicted and true surface normal maps measured in degrees ($0$ to $180$). In our evaluation, we apply each algorithm to PS-Wild test dataset with three different lighting methods and discuss the result mainly based on the averaged MAE over 50 different objects. 
\subsection{Ablation study}
\label{sec:ablation}
\noindent\textbf{Base architecture:} As needed, we define the base architecture for the ablation study with following design elements: SwinTransformer~\cite{Liu2021Swin} for feature extraction; No communication in the encoder; $256\times 256$ canonical resolution; Three stacks of transformer layers followed by PMA for the feature aggregation in the decoder; In ablation, only the target property was changed from this base architecture.
\vspace{4.0pt}\\
\noindent\textbf{Encoder}: \Tref{tab:enc} shows the comparison of four different image feature extractors: SwinTransformer~\cite{Liu2021Swin}, VisionLongformer~\cite{Zhang2021} and ResNet-50/101~\cite{He2016}. We observe that Transformer-based encoders, especially SwinTransformer outperformed ResNets probably due to the larger receptive field of the Transformer model. Because the main purpose of this work is to present a viable method to demonstrate our ideas, the further discussion remains future work.
\begin{table*}[t]
\begin{minipage}[t]{80mm}
\newcolumntype{C}{>{\centering\arraybackslash}X}
\newcolumntype{L}{>{\raggedright\arraybackslash}X}
\setlength{\tabcolsep}{1mm} 
    \centering
    \vspace{-5pt}
    \caption{Different number of input images (Ours, pre-fusion).}
    \vspace{-10pt}
    \small
    {\renewcommand\arraystretch{0.7}
    \begin{tabularx}{80mm}{LCCCCC}
    \toprule
        & 1 & 4 & 8 & 16 & 32\\
         \midrule
        Directional & 35.4 & 22.6 & 18.8 & 17.5 & 17.0\\
        HDRI & 27.2 & 19.5 & 16.5 & 14.7 & 14.5\\
        Dir.+HDRI & 28.5 & 20.6 & 17.4 & 15.6 & 13.8\\
    \bottomrule
    \end{tabularx}
    }
    \label{tab:num}
\end{minipage}
\hspace{10mm}
\begin{minipage}[t]{80mm}
\newcolumntype{C}{>{\centering\arraybackslash}X}
\setlength{\tabcolsep}{1mm} 
    \centering
    \vspace{-5pt}
    \caption{Comparison with other methods (pre-fusion).}
    \vspace{-10pt}
    \small
    {\renewcommand\arraystretch{0.7}
    \begin{tabularx}{80mm}{XCCCC}
    \toprule
         & Ours & GCNet~\cite{Chen2019} & MPM~\cite{Mo2018} & Var.~\cite{Haefner2019} \\
        \midrule
        Directional & \textbf{17.0} & 17.7 & 32.9 & 33.0 \\
        HDRI & \textbf{14.5} & 24.8 & 30.4 & 37.5 \\
        Dir.+HDRI & \textbf{13.8} & 31.5 & 30.8 & 32.5 \\
    \bottomrule
    \end{tabularx}
    }
    \label{tab:comp}
\end{minipage}
\end{table*}
\begin{figure*}[!t]
	\begin{center}
		\includegraphics[width=175mm]{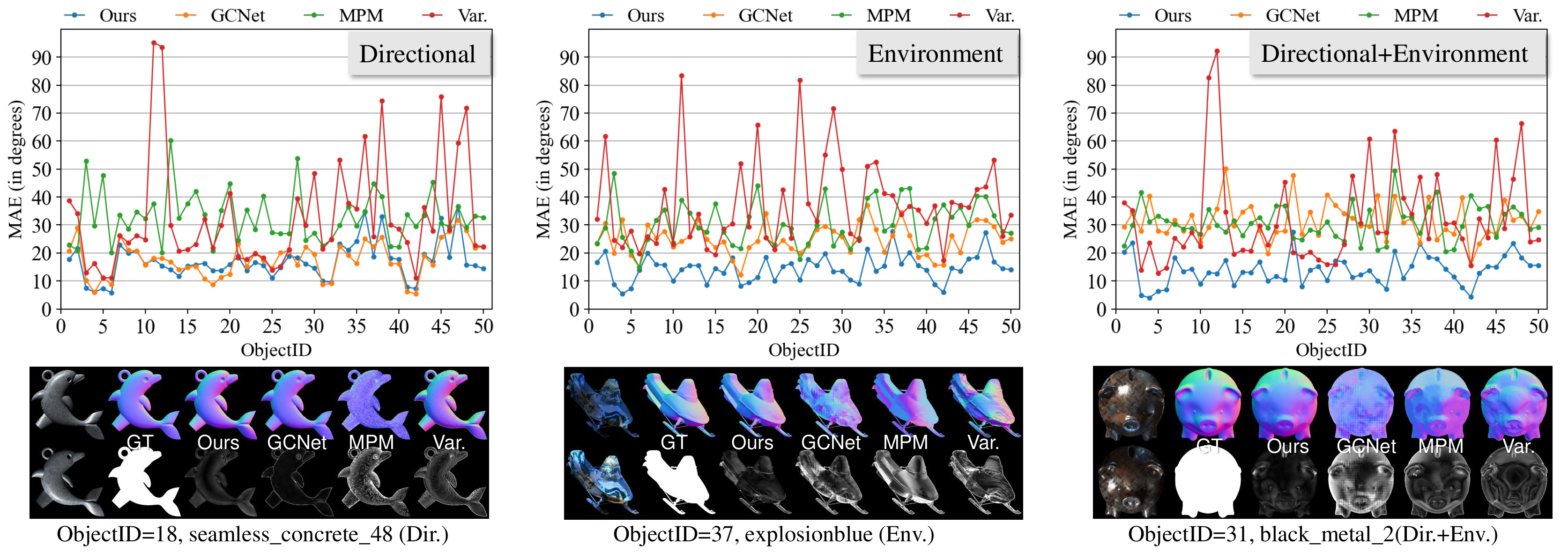}
	\end{center}
	\vspace{-15pt}
	\caption{We compared our method with three uncalibrated photometric stereo methods specifically designed for single directional lighting (\ie GCNet~\cite{Chen2019}) and natural lighting (\ie MPM~\cite{Mo2018} and Variational~\cite{Haefner2019}) on our synthetic PS-Wild test dataset. There were three different methods of lighting; directional, HDRI ( and the mixture of them. The performance was evaluated based on the mean angular errors in degrees. We showed the examples in the dataset and the recovered surface normal maps/error maps (80 degrees at maximum). }
	\label{fig:test_comp}
	\vspace{-5pt}
\end{figure*}

\Tref{tab:interaction} compared four different variants of the feature communication; no communication in the encoder (Base) during the image feature extraction (Dur-Ext), before the multi-scale fusion (Pre-Fus) and after the multi-scale fusion (Post-Fus) as illustrated in~\Fref{fig:interaction}. As have been mentioned already, the best performance was obtained when the communication was done just before the multi-scale fusion. Interestingly, the second best was obtained when no communication among images was done in the encoder, but only in the decoder. We analysed this result and found that the communication during the image feature extraction simply broke the feature extraction process and the post-fusion communication made the optimization unstable since it is equivalent to increasing the transformer layers in the decoding process which is consistent with later results. 

\Tref{tab:cano} compared different canonical resolutions either from 128$\times$128 or 256$\times$256 and we observe that 256$\times$256 worked slightly better. This result indicates that though the lower resolution is helpful to see the entire object, important information for details could be discarded at very low resolution. On the other hand, it is surprising to see that 32$\times$32 global lighting contexts from 128$\times$128 canonical resolution still provides a reasonable reconstruction for 512$\times$512 output resolution. Unfortunately, we couldn't get the result of 512$\times$512 canonical resolution due to the memory limit. 

Finally, we compared our spatially-varying global lighting contexts with the spatially-{\it uniform} ones by applying the global average pooling to shrink $\{\mathcal{G}\}\in \mathbb{R}^{s\times s\times d_e}$ to $\{\mathcal{G}_{uni}\}\in \mathbb{R}^{1\times1\times d_e}$ and feeding the spatially same vector to the decoder. In reality, this is a similar procedure in existing uncalibrated photometric stereo networks~\cite{Chen2019,Kaya2021} where the single lighting parameter is firstly recovered for each image and used as the input of the surface normal predictor. However, \Tref{tab:global} shows that the network with the uniform context always provided the trivial solution since it couldn't capture spatially-varying lighting effects.  
\vspace{4.0pt}\\
\noindent\textbf{Decoder:}
We compared three different variants of the pixelwise feature aggregation strategy from a max pooling, three Transformer layers with PMA and six Transformer layers with PMA; The further to the right we go, the more complex interaction can theoretically take place. Note that the latent vector size in the Transformer layer was fixed by $384$ and the hidden dimension of the feedforward layer was $1024$. The result is illustrated in~\Tref{tab:dec}. Though there is no surprise to see the simple max-pooling didn't work at all, the reason for the worse results for the deeper Transformer networks seemed to be due to the over-fitting. It should be noted that the Transformer network is known to be a difficult network to train, and careful adjustment of the hyperparameters may yield different results.
\vspace{4.0pt}\\
\noindent\textbf{Different number of input images:}
In \Tref{tab:num}, we evaluated our network (base+pre-fusion) with different number of input images. As expected, the accuracy dropped as the number of input images decreased. However, unlike exiting calibrated and uncalibrated photometric stereo methods which had been shown that they didn't work when the number of images was small (\eg 10)~\cite{Ikehata2018,Chen2019}, the degree of decline with our method was not quite significant. 
\subsection{Quantitative Evaluation on PS-Wild Test Data}
\begin{figure*}[!t]
	\begin{center}
		\includegraphics[width=170mm]{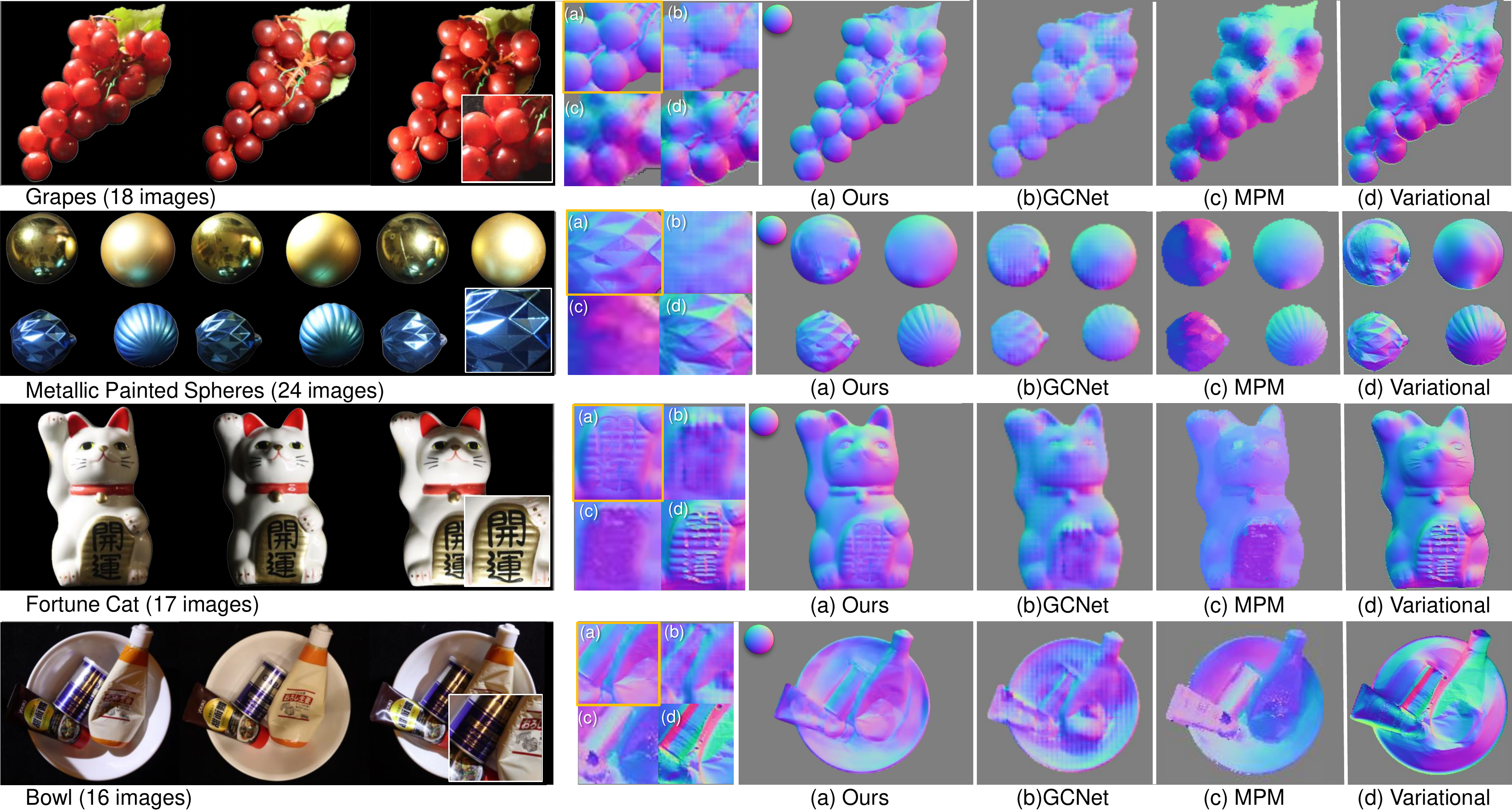}
	\end{center}
	\vspace{-15pt}
	\caption{The quantitative comparison on real images under challenging spatially-varying lighting conditions. All the images were captured with a moving near area light source and static indoor natural illumination. Foreground objects have been cropped off using object masks.}
	\vspace{-7pt}
	\label{fig:real_comp}
\end{figure*}
We compared our method to state-of-the-arts on the uncalibrated photometric stereo problem on our synthetic test dataset. The first comparison point is GCNet~\cite{Chen2020} which is a learning-based method under the directional lighting condition that alternately estimates physical lighting parameters and surface normals. The second point is the matrix-based patch merging (MPM)~\cite{Mo2018} which is the Lambertian photometric stereo method for uncalibrated natural lighting based on the equivalent directional lighting model. The third point is the variational uncalibrated photometric stereo method (Variational)~\cite{Haefner2019} which approximates Lambertian reflectance model through a spherical harmonic expansion. We used authors' official codes for the evaluation and our GLC-UniPS architecture was ``base + pre-fusion'' configuration that has yielded the best result so far. 

The results are illustrated in \Fref{fig:test_comp} and \Tref{tab:comp}\footnote{The input images and recovered normal map for individual objects are presented in the supplementary.}. We observe that our UniPS-GLC reasonably worked independent of the lighting methods. As expected, GCNet~\cite{Chen2019} worked fine for objects under the directional lighting condition (\eg ID 48), however problematic with non-uniform materials and non-directional lightings. Overall, MPM~\cite{Mo2018} and Variational~\cite{Haefner2019} were inferior to GCNet for diverse objects due to their assumptions of convex, Lambertian surfaces. Especially, MPM had a problem in handling spatially varying non-Lambertian textures or non-convex geometries (\eg ID 37) and variational also had a problem in handling non-Lambertian reflections especially for metallic surfaces (\eg ID 31). Interestingly, better performance of our method was observed for objects of the same shapes and materials under the environment lighting than the directional one. This result may be attributed to the fact that in an image under the directional lighting condition, the shadow or inter-reflection behaves negatively, whereas under the environment light, they positively yields spatially-varying lighting effects as a source of information.


\subsection{Qualitative Evaluation on Real Data}\label{sec:realdata}
\Fref{fig:realdata} validates our method (base+pre-fusion) on real images. Objects of varying materials such as ceramic, metallic and  clear coat were captured by a 8-bit smartphone camera under a few hand-held area lights and/or the indoor natural illumination. The area light sources were intentionally put within 30 cm of an object so that strong spatially-varying lighting effects were observed. Although quantitative evaluation is not possible due to the absence of the ground truth, we observe that overall reasonable normal maps were recovered even under these extremely challenging setups. We also observe that the very low resolution of GLiCo (\ie 64$\times$ 64) is sufficient to recover surface details.

For a more objective evaluation, \Fref{fig:real_comp} provides a comparison with GCNet, MPM and Variational. Images were captured by putting a single area light near an object under the static indoor natural illumination; therefore, could not be represented by physical lighting models such as the directional lighting nor spherical harmonics. We picked four objects with different difficulties and included two ``spherical'' objects (\ie {\it Grapes} and {\it Metallic Painted Spheres}) so that the quality of normal maps could be evaluated based on their spherical parts. {\it Grapes} exhibit complex cast shadows because of its non-convex geometry. {\it Metallic Painted Spheres} consists of four metallic-painted balls with different surface roughness and structures which exhibit strong inter-reflections with each other. {\it Fortune Cat} contains large ``black'' regions, which are known to be difficult to distinguish from shadows. Bowl includes multiple challenging objects put in a bowl which is hard to resolve the convex-concave ambiguity. We emphasize that these are extremely difficult objects that existing photometric stereos methods have not even attempted to handle them. Though our results are still far from perfect (\ie, we clearly observe errors in our results due to inter-reflection in {\it Metallic Painted Spheres} and convex geometries in {\it Bowl}), our method obviously recovered most reasonable surface normal maps.
\section{Conclusion}
This paper tackled a new photometric stereo task, named universal photometric stereo (UniPS) which drops prior assumptions of physical lighting models. To this end, we presented the first viable UniPS network based on the generic lighting representation named global lighting contexts. We also presented synthetic training and evaluation datasets for our UniPS task and our extensive evaluation demonstrated the performance of our method. 

Despite the significant progress toward a practical photometric stereo method, there are some limitations remained. First, the deeper analysis of global lighting contexts is missing. To explore its potential for applications other than the surface normal recovery is a future work. Second, our training method specifies a camera model (\eg, orthographic, linear camera) and we may also want to remove this dependence. Finally, we need a real dataset with the ground truth for evaluating UniPS methods in a more quantitative manner.
\section*{Aknowledgement}
This work was supported by JSPS KAKENHI Grant Number JP22K17919.
\\
{\small
\bibliographystyle{ieee_fullname}
\bibliography{egbib}
}
\newpage
\section*{Appendix}
\subsection*{Appendix A. Implementation Details}
\label{sec:details}
\noindent\textbf{Architecture details:} In the main paper, an overview of our universal photometric stereo network was given but some important details were omitted due to the space limit. In this section, we detail the ``basic + pre-fusion'' configuration of our universal photometric stereo network. It should be noted that the hyper parameters in our architecture are all selected empirically, so it is quite possible that there are parameters that will give better performance.

The image-wise feature extraction network (\ie Swin-S variant of SwinTransformer~\cite{Liu2021Swin}) and subsequent multi-scale feature fusion with the feature pyramid network (\ie UPerNet~\cite{Xiao2018}) in our encoder were implemented on MMSegmentation~\cite{MMSegmentation}. The updates from the original codes are mainly two. First, we input a mask image in addition to an RGB image. Second, following the suggestions in~\cite{Xiao2021}, we modified the original mlp-based patch embedding (\ie, local information embedding during the reduction of the image resolution to 1/4 of the canonical resolution) to the CNN-based one with five convolutional layers to capture the local shading variations. The number of different scales was four; hence, given the canonical resolution of 256$\times$256, sizes of the multi-scale feature maps were 64$\times$64$\times$96, 32$\times$32$\times$192, 16$\times$16$\times$384 and 8$\times$8$\times$768, which were fused to 64$\times$64$\times$256 global lighting contexts.

\begin{figure*}[t]
	\begin{center}
		\includegraphics[width=160mm]{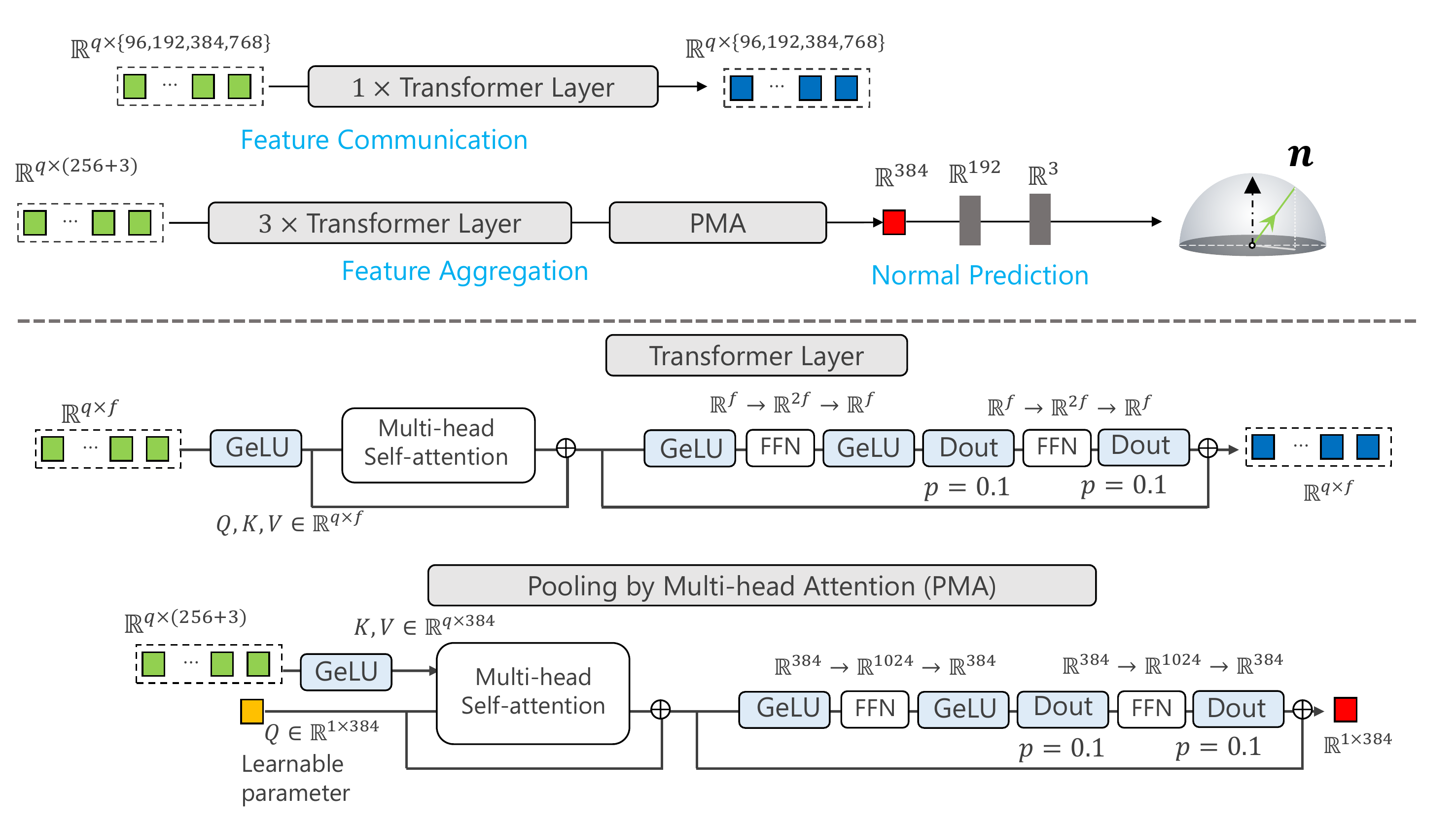}
	\end{center}
	\caption{Implementation details of the feature communication and aggregation steps. Our feature communication step is composed of a single Transformer layer and our feature aggregation step is composed of three Transformer layers followed by PMA layer.}
	\label{fig:arch_detail}
\end{figure*}

The feature communication in our encoder and aggregation in our decoder were pixelwisely applied to feature vectors under different lighting conditions in similar to our previous work~\cite{Ikehata2021}. As illustrated in~\Fref{fig:arch_detail}, the feature communication step built upon a single Transformer layer where input feature vectors were firstly projected to query, key and value vectors whose dimensions were same with the input ones. They were then passed to a multi-head self-attention (the number of heads is $8$) and a multi-layer perceptron (MLP) with the pre-layer normalization~\cite{Xiong2020} and dropout ($p=0.1$). Though the MLP doubled the original feature dimension, the feature dimension and number of feature vectors in a set did not change between the input and output of the feature communication step. 

The feature aggregation step input $q$ sets of vectors ${\rm cat}\{I(\bm{x}), \mathcal{G}(\mathcal{S}(\bm{x}))\}_{\{1\dots q\}}\in \mathbb{R}^{q\times(256+3)}$ where each vector was composed of raw pixel values and the interpolated global lighting context. Then the input set was passed to three Transformer layers and a PMA~\cite{Lee2019} where the number of elements in a set was shrunk from $q$ to one. The surface normal predictor was a MLP with one hidden layer whose feature dimension shrank as $384\rightarrow 192 \rightarrow 3$ and the norm of the output vector was normalized to be a unit surface normal vector at the location.
\vspace{4.0pt}\\
\noindent\textbf{Competitor details:} It should be noted again that all the algorithms (ours, GCNet~\cite{Chen2020}, MPM~\cite{Mo2018} and Variational~\cite{Haefner2019}) took the object mask as input. To ensure a fair comparison, we applied the same center crop to input images, which means that the input of all the algorithms were exactly same (\ie, crops of images and an object mask). For a fair evaluation, we used the authors' official implementations for competitors. Since there is a binary ambiguity left in the surface normal recovered by MPM (\ie signs of $x,y,z$ directions), we manually solved it so as to be quantitatively optimal in the quantitative experiments and most visually plausible in the qualitative evaluation. As for GCNet, we used the pretrained model provided by authors since our training dataset was not available for their model due to the fact that GCNet requires the supervision of directional lightings. In addition, we found that GCNet~\cite{Chen2019} didn't work at all for our raw test images without the proper image normalization (The data normalization is also important for the DiLiGenT~\cite{Shi2018} evaluation), therefore we empirically performed the linear image normalization dividing each image by $0.1\cdot {\rm max}(I)$ so that the pixel values in each image ranged between $0$ and $0.1$. Unlike others, Variational~\cite{Haefner2019} is actually an algorithm for perspective images and it requires the focal length of images as input. So we approximated test images as perspective ones by using the unit focal length (\ie $f=256$ for 256$\times$256 image) for our PS-Wild test dataset and using ones from Exif-Tags in the real evaluation. We note that empirically, the small differences of the focal length didn't show any significant difference in results. Unfortunately, MPM~\cite{Mo2018} is a quite computationally expensive algorithm whose computational complexity is $O(h^2w^2)$ and we confirmed that it didn't work for images whose sizes were bigger than 512$\times$512. For a fair comparison, we used 256$\times$256 crops in both quantitative and qualitative comparison because we confirmed that MAE didn't significantly depend on the input image resolution.
\subsection*{Appendix B. PS-Wild Dataset and Training Details}
\begin{figure*}[t]
	\begin{center}
		\includegraphics[width=160mm]{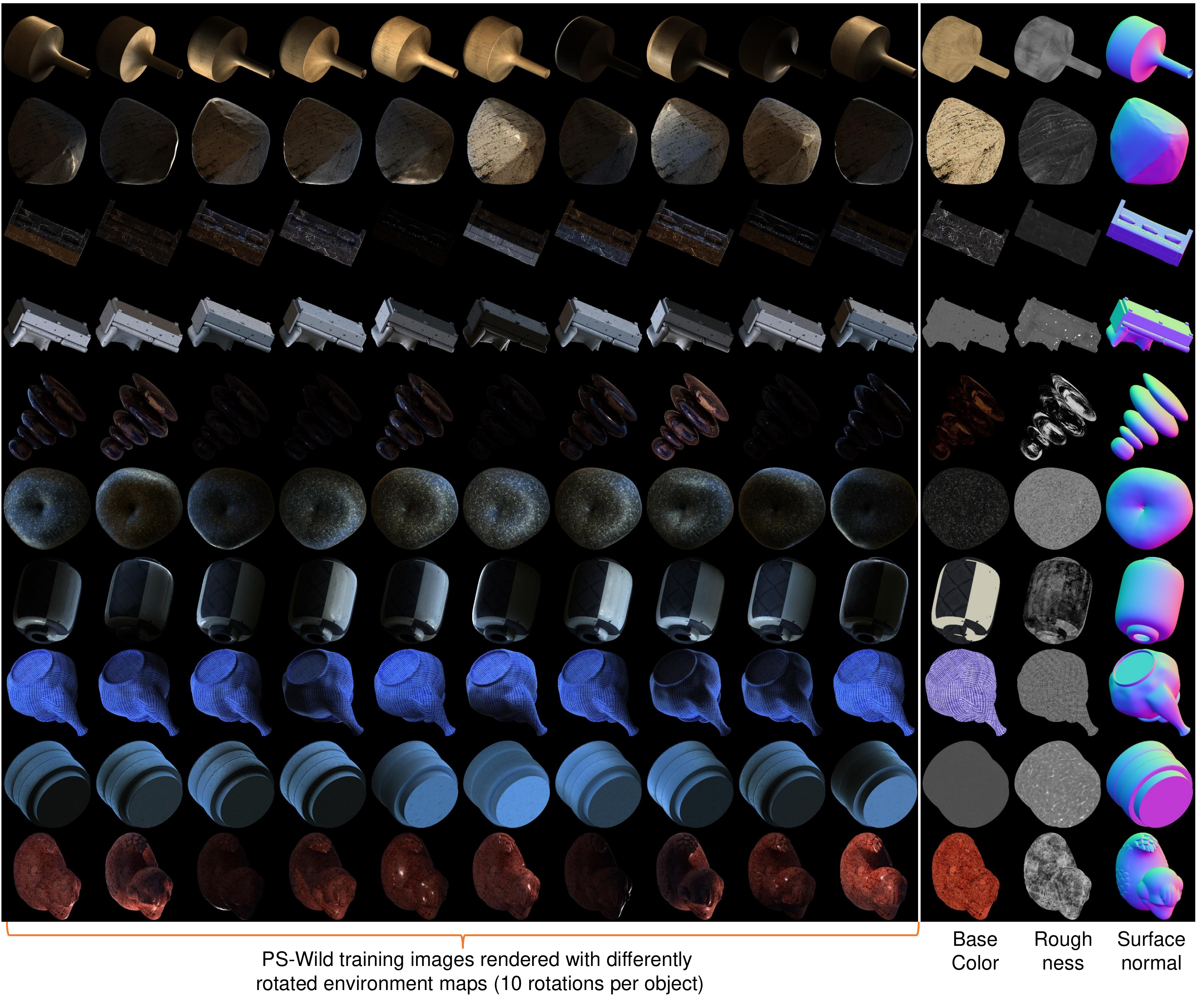}
	\end{center}
	\vspace{-15pt}
	\caption{Examples of images and BRDF parameter maps in our PS-Wild training dataset (metallic map is omitted).}
	\label{fig:pswild_train}
\end{figure*}
\noindent\textbf{Renderer details:} 
PS-Wild was rendered with the {\it Cycles} engine in Blender2.93~\cite{Blender}. For a full global illumination rendering using a path tracing integrator with direct light sampling, we used 256 rendering samples with 10 max ray bounces. Each BRDF material in both training and test 3-D assets consisted of 2-D texture maps of the base color, roughness, metalness which were directly fed to the diffuse and specular BRDFs of the Cycles engine (\ie we used Principled BSDF shader~\cite{DisneyPrincipledBSDF}). In~\Fref{fig:pswild_train}, we illustrated some examples of rendered images and 3-D assets in our PS-Wild training dataset. Each row corresponds to one object from 10,099 objects in total. As for the test dataset, we illustrate the entire 50 objects and corresponding results in Fig.3-52. As mentioned, our textures are classified into three types of materials (six as categorized in ShareTextures~\cite{ShareTextures}); diffuse (Fabric, Concrete), specular (Wood, Floor, Ground) and metallic (Metal). The rendering pipeline was exactly same as one for the training dataset.
\\\\
\noindent\textbf{Training details:} We augmented the dataset during the training to bring more variations in training examples. Concretely, we randomly flipped images horizontally or vertically, and randomly rotated images by $90$ degrees. In addition, we also performed the random color swapping for each image since our task didn't include the surface reflectance recovery. We used $p=0.5$ for all the augmentations.  

\end{document}